# Ordinal Pooling Networks: For Preserving Information over Shrinking Feature Maps

Ashwani Kumar[1]

Department of Electronic and Electrical Engineering
University of Sheffield
George Porter Building, Broad Lane, Sheffield S3 7HQ, United Kingdom
akumar4@sheffield.ac.uk

## Abstract

In the framework of convolutional neural networks that lie at the heart of deep learning, downsampling is often performed with a max-pooling operation that only retains the element with maximum activation, while completely discarding the information contained in other elements in a pooling region. To address this issue, a novel pooling scheme, Ordinal Pooling Network (OPN), is introduced in this work. OPN rearranges all the elements of a pooling region in a sequence and assigns different weights to these elements based upon their orders in the sequence, where the weights are learned via the gradient-based optimisation. The results of our small-scale experiments on image classification task demonstrate that this scheme leads to a consistent improvement in the accuracy over max-pooling operation. This improvement is expected to increase in deeper networks, where several layers of pooling become necessary.

## 1. Introduction

Convolutional neural networks (CNNs) [1] that are specifically suited for various visual tasks, such as image classification [2], object detection [3], segmentation [4], and modelling video evolution [5], are one of the main drivers of deep learning. CNNs utilize a miniature neural network, referred as feature detector, which is replicated over space to extract various features from an image. Thus, CNNs employ a much smaller set of parameters in comparison to the fully connected neural networks (FCNNs) for the layers of equivalent spatial size, which allows them to scale up to high-resolution images. A typical deep CNN architecture consists of 3 types of layers: 1) convolutional: for extracting various features or activations from an input image or a feature map, 2) pooling: a down-sampling technique for aggregating the activations of the neighbouring elements within a pooling region so that the size of the feature map along the spatial dimensions becomes smaller, and 3) FCNN: to carry out the classification from the extracted features at the end of the network, which can be either linear or non-linear. Many variants of CNNs have been reported in literature, for instance, network-in-network (NIN) [6], residual networks (ResNets) [7], inception-v4 [8], squeeze-and-excitation networks (SENets) [9], and dense convolutional networks (DenseNets) [10].

Replicating the feature detector across the spatial dimensions in CNNs enables the weight sharing across the space. This helps achieve equivariance, i.e. a translation of an object in an input image

---

results in an equivalent translation in the activations at the output feature map. The pooling operation, on the other hand, tends to achieve translational invariance, which is most commonly obtained either by avg-pooling, where all the activations in a pooling region are averaged together, or by max-pooling, where only the element with the maximum activation is retained. A theoretical analysis on these two pooling operations reveals that none of the techniques is optimal [11]. Yet, it is observed that max-pooling in general achieves better performance over avg-pooling [12], [13]. This is due to the fact that avg-pooling treats all the elements equivalently irrespective of their activations, which results in an undervaluation of the elements with higher activations, while the elements with smaller activations are overestimated.

This work proposes an alternate pooling scheme: *Ordinal Pooling Network* (OPN) that resolves this issue of unfair valuation of the elements in a pooling region, while still preserving the information from other activations. In this scheme, all the elements in a pooling region are first ordered in a sequence based upon their activations and then combined together via a weighted sum, where the weights are assigned depending upon the orders or ranks of the elements and are learned during the gradient-based optimization or training. The idea of a rank-based weight aggregation was first introduced by A. Kolesnikov *et al.* [14], who propose a global weighted rank-pooling (GWRP) in order to estimate a score associated with a segmentation class, in the context of image segmentation. Unlike the regular pooling, which performs pooling operation only upon the set of elements inside a pooling region, GWRP acts upon all the elements in a feature map to generate the score of a particular segmentation class. In GWRP, all the elements of a feature map are first sorted in the descending order, depending upon their scores for a particular segmentation class, which is similar to our case. However, the weights that are assigned based on the order of the elements are determined from a hyperparameter and therefore do not change during the training.

A key difference between OPN and a pooling operation is that while a typical pooling acts upon one feature map at a time, OPN consists of a different set of weights for each feature map and therefore pools features from all the feature maps simultaneously. Owing to this fact, OPN is referred in this work as a *pooling network* rather than a *pooling operation*.

## 2. Related Works

Since a 2 × 2 pooling with a stride of 2 discards 75% of the feature map upon its application, it is an aggressive operation, which after a series of applications can result in a significant loss in information. To apply pooling in a gentler manner, a fractional max-pooling [16] has been proposed, where the dimensions of the feature map can be reduced by a non-integer factor. On the other hand, for allowing information from other activations within a pooling region to also pass to the next stage, a stochastic version of pooling has been proposed by M. Zeiler *et al.* [17], where an element in a pooling region is selected based upon its probability within the multinomial distribution constructed from all the activations inside a pooling region. Another stochastic variant of pooling, S3Pool [18], is a 2-step pooling technique, where in the 1$^{st}$ step, a 2 × 2 pooling with a stride of 1 is applied, while in the 2$^{nd}$ step, a stochastic downsampling is performed. A combination of these operations makes S3Pool to work as a strong regularisation.

The other works focus upon generalising the pooling operation. C. Gulcehre *et al.* [19] regard the pooling as an *lp* norm operation, where the values of 1 and ∞ for the parameter *p* correspond to avg-pooling and max-pooling, whereas $p$ itself is learned during the training. C.-Y. Lee *et al.* [20] propose mixing together avg- and max-pooling operations by a trainable parameter and also introduce the idea of tree pooling, which is a way to learn different pooling filters and also learn to responsively combine these filters.

In all convolutional net (All-CNN) [21], the pooling is replaced with another convolutional layer of equivalent stride and filter size. Another approach, RNN-based pooling [22], replaces the pooling operation with a long short-term memory (LSTM) unit, which is a variant of recurrent neural network (RNN). In this case, after all the activations in each pooling region have been scanned sequentially, the final output from RNN layer is returned as the pooled value.

In addition to GWRP [14], another variants of OPN, have been introduced by Z. Shi *et al.* [15], who propose three pooling schemes based upon the rank of the elements: 1) average, 2) weighted, and 3) stochastic. Unlike OPN, all of these schemes require an additional hyperparameter that, in the first scheme, is used to determine the threshold for choosing the activations to be averaged. In the second scheme, the hyperparameter is used to generate the weights to be applied to the activations, which remain fixed across all the feature maps, while in the third scheme, a set of probabilities are generated based upon this hyperparameter that is used to select an element in a pooling region.

To completely circumvent the issue of loss of information via pooling, recently a novel architecture, CapsNet [23], is proposed. CapsNet consists of layers of capsules, where each capsule represents a group of neurons. The outputs from the capsules in one layer are routed to the capsules in the subsequent layer based upon the assignment coefficients which are determined from the Expectation-Maximisation algorithm. In comparison, the ordinal pooling networks proposed in this work are compatible with the existing CNN framework, and can prove to be an effective replacement to the conventional pooling operations in the CNN based architectures that are currently in operation.

## 3. Ordinal Pooling Networks

In general, a pooling operation $f_P$ can be defined as

$$s_j = f_P(a_{ij}) \quad \forall i \in R_P \ \& \ \forall j \in [1, N] \tag{1}$$

where $R_P$ represents the pooling region in a feature map indexed as $j$, $N$ indicates the total number of feature maps or channels, and $a_{ij}$ denotes all the activations within $R_P$. For avg-pooling, Eq. (1) can be expressed as

$$s_j = \frac{1}{|R_P|} \sum_{i \in R_P} a_{ij} \quad \forall j \in [1, N] \tag{2}$$

where $|R_P|$ represents the number of elements in $R_P$, which is assumed to be constant across all the channels. For max-pooling, Eq. (1) results in

$$s_j = \max(a_{ij}) \quad \forall i \in R_P \ \& \ \forall j \in [1, N] \tag{3}$$

where max() returns a maximum of all the activations in $R_P$. Since max-pooling only passes the elements with a maximum activation, it losses too much information, as the size of the feature map along the spatial dimensions is reduced. Therefore, in the most recent CNN architectures, such as in ResNets or DenseNets, avg-pooling has been adopted to perform the downsampling operation. An alternative to avg- or max- pooling schemes would be to take a weighted combination of different activations in a pooling region, as

$$s_j = \sum_{i \in R_P} w_{ij} a_{ij} \quad \forall j \in [1, N] \qquad (4)$$

This scheme is referred here as location-based pooling network (LBPN), since each feature map $j$ has its own set of weights $w_{ij}$, while the weights are assigned based upon the location $i$ of the activations in $R_P$, and learned from the gradient-based training. Notice that this scheme is described as a pooling network, since it employs a different set of weights for each of the feature maps. This is similar to replacing a pooling operation with a convolutional neural network, such as in All-CNN, except that the total number of parameters in this operation is $|R_P| \times N$. Whereas, an equivalent convolution operation with a filter size $|R_C|$ along one feature map, requires a total of $|R_C| \times N^2 + N$ parameters (where $N$ in the summation accounts for the bias terms), since in this case, features from different channels are also allowed to mix together. However, even with the trainable weights, LBPN will not be effective in properly weighting the different activations, since the distribution of activations in the pooling region does not remain fixed but varies as the pooling filter scans different parts of a feature map.

This problem can be overcome by the proposed ordinal pooling networks, where the elements in a pooling region are first sorted based upon their activations and then combined together by a weighted sum, while the weights are assigned based upon the order of the elements in the sequence and learned during the gradient-based training. This operation can be written as:

$$s_j = \sum_{i \in R_P} w\left(ord(a_{ij})\right) a_{ij} \quad \forall j \in [1, N] \qquad (5)$$

where a weight is determined from the order, $ord()$, of the activation. This scheme makes the weights invariant to the activation. Therefore, an element within a pooling region is always assigned the same weight, as long as its order in the sorted sequence remains unchanged. Both LBPN and OPN employ the same number of parameters, which, unlike a convolutional layer, is linear in $N$.

The additional sorting step in OPN is essential in introducing the nonlinearity, which is also present in max-pooling or $lp$-norm pooling and therefore, sets the operation of OPN apart from LBPN or avg-pooling, which are linear. On account of this nonlinearity, OPN can generalise both avg-pooling and max-pooling. For instance, all the weights equal to $1/|R_P|$ result in the avg-pooling, while a weight corresponding to the maximum activation equal to 1 and rest of the weights equal to 0 represent the max-pooling. The sorting step enables OPNs to overcome the problem of under- or over- valuation of larger or smaller activation, and therefore facilitates preserving of information from other activations in an efficient manner.

For an exemplary $2 \times 2$ region, as shown in Fig. 1 (a), different pooling operations, represented by Eqs. (2)—(5), are illustrated in Figs. 1 (b)—1 (e), respectively.

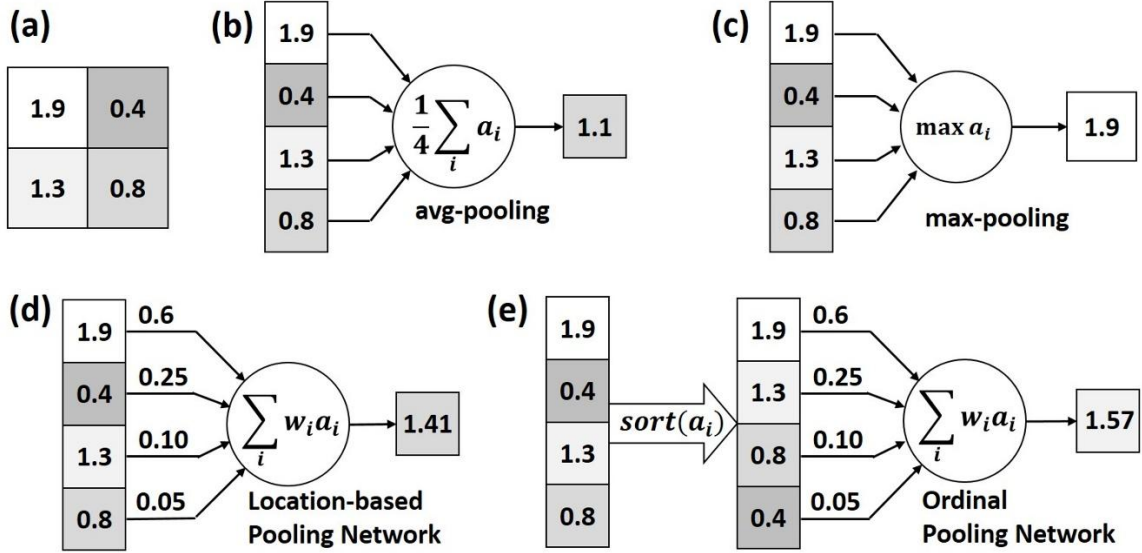

Fig. 1. Toy examples of various pooling operations. (a) 2x2 pooling region, (b) avg-pooling, (c) max-pooling, (d) location-based pooling network, and (e) Ordinal pooling network.

## 4. Experiments

We employ a very simple architecture of convolutional neural networks and test its performance for the classification task on MNIST database of handwritten digits [24]. As shown in Fig. 2, this network consists of two stages of alternating convolutional and pooling layers, which are followed by two layers of fully-connected neural networks. Rectified Linear Units (ReLUs) are utilised as the activation functions after each convolutional and hidden layers. Training is performed using stochastic gradient descent (SGD) with a dropout regularisation [25]. In the dropout, the probability of keeping the connections is started from 98% and decayed exponentially at a rate of ~0.003 for every 1000 batches of training, which is found to be helpful in improving the convergence at the beginning of the training. More details upon the implementation can be accessed by following the code at: https://github.com/ash80/Ordinal-Pooling-Networks, implemented with TensorFlow [26].

The following three pooling schemes are considered for all the pooling layers (filter size=$2 \times 2$, stride=2) in Fig. 2:

1) **Max Pooling:** To form a baseline for the results.
2) **Location-based Pooling Nets (LBPNs):** pooling networks defined by Eq. (4), where trainable weights are assigned based upon the location of the activations (cf. Fig. 1 (d)), while a different set of weights is considered for each individual channel. Hence, for a total number of $N$ channels and a pooling region with a size of $2 \times 2$ per channel, LBPNs require a total of $4N$ parameters per pooling layer. LBPNs are needed to ensure that any improvement in the performance for OPNs is not simply the result of the increase in the number of parameters when different pooling schemes are compared.
3) **Ordinal Pooling Nets (OPNs):** pooling networks defined by Eq. (5), where weights are assigned based upon the sorted sequence of activations within a pooling region. Similar to LBPNs, OPNs also increase the total number of parameters by $4N$ per pooling layer.

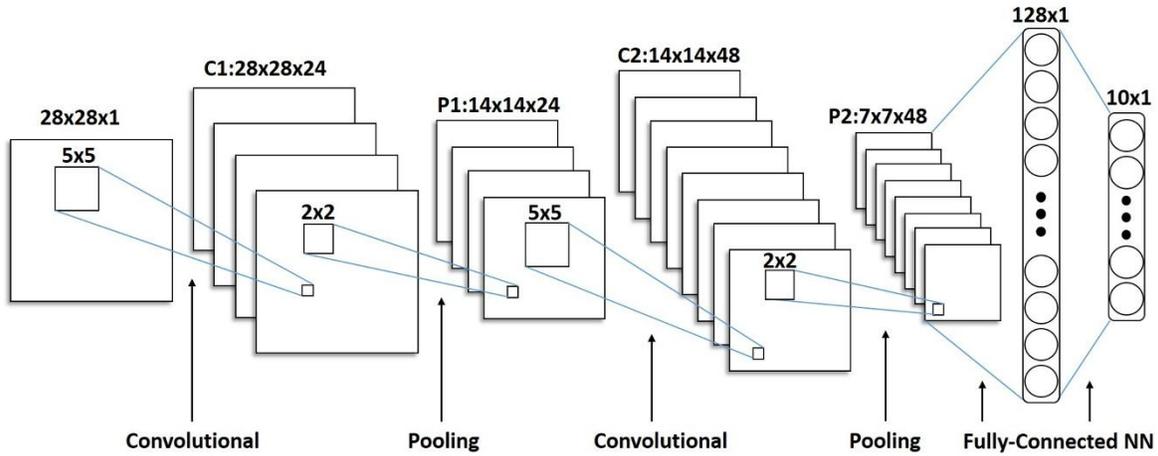

Fig. 2. The architecture of the convolutional neural network employed in this work for the classification of digits in MNIST database.

## 5. Results

The results of the performance of the pooling schemes defined in 1)—3), are compared in Fig. 3. As can be observed, replacing the max-pooling with OPNs leads to a validation accuracy, which on average remains ~0.10% higher (10 images over a validation set of 10000 images) than that achieved by either max-pooling or LBPNs.

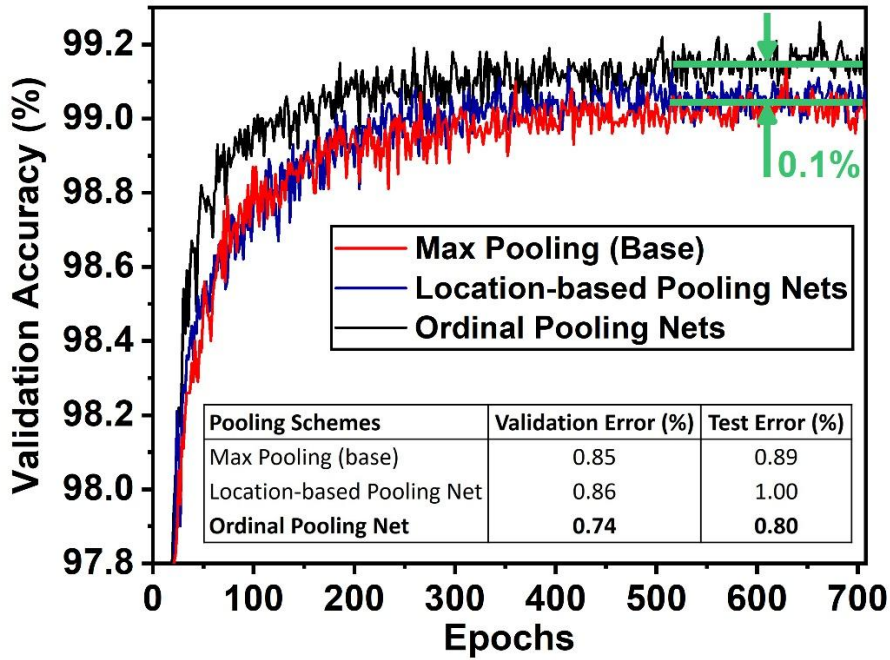

Fig. 3. Validation accuracy of the three pooling schemes vs. the number of epochs during the training. The table provided in the inset highlights the minimum error during the validation and test times.

One caveat about OPNs is that they tend to be sensitive to the initial values of the weights in OPN. In general, initialising the weights such that all the weights are positive, sum to 1, and are greater for the higher activations works best. An example of such weights is shown in Fig. 1 (e) for a $2 \times 2$ pooling region. We also experimented with a variant of OPN, where all the weights were restricted to be positive and normalised to 1 with a softmax function (not shown here). This, however, resulted in a marginal degradation in the performance compared to the results for OPNs in Fig. 3, where apart from a proper initialisation no restriction is imposed upon the weights.

Figs. 4 (a) and 4 (b) show the heat maps of the weights corresponding to the two layers of OPNs in Fig. 2, before and after the training. Before the training, the weights for each of the channels are initialised to the same set of values. However, as can be seen from the figures, the final values of the weights for every channel converge to a different set of values after the training and no longer remain normalised. At these weights, every feature map at the output of the pooling layer retains the maximum information from its corresponding input feature map that is required to minimise the classification error. This fine-tuning of the weights to the data at a feature map level is believed to be the cause of the faster convergence of OPNs seen in Fig. 3 compared to max-pooling or LBPNs.

It can also be observed from Figs. 4 (a) and (b) that the final value of a weight with smaller index, which is assigned to the elements with higher activation, remains greater than the weights with higher indices, assigned to the elements with lower activations, along all the channels, thereby suggesting that elements with higher activations are valued more than those with smaller activations.

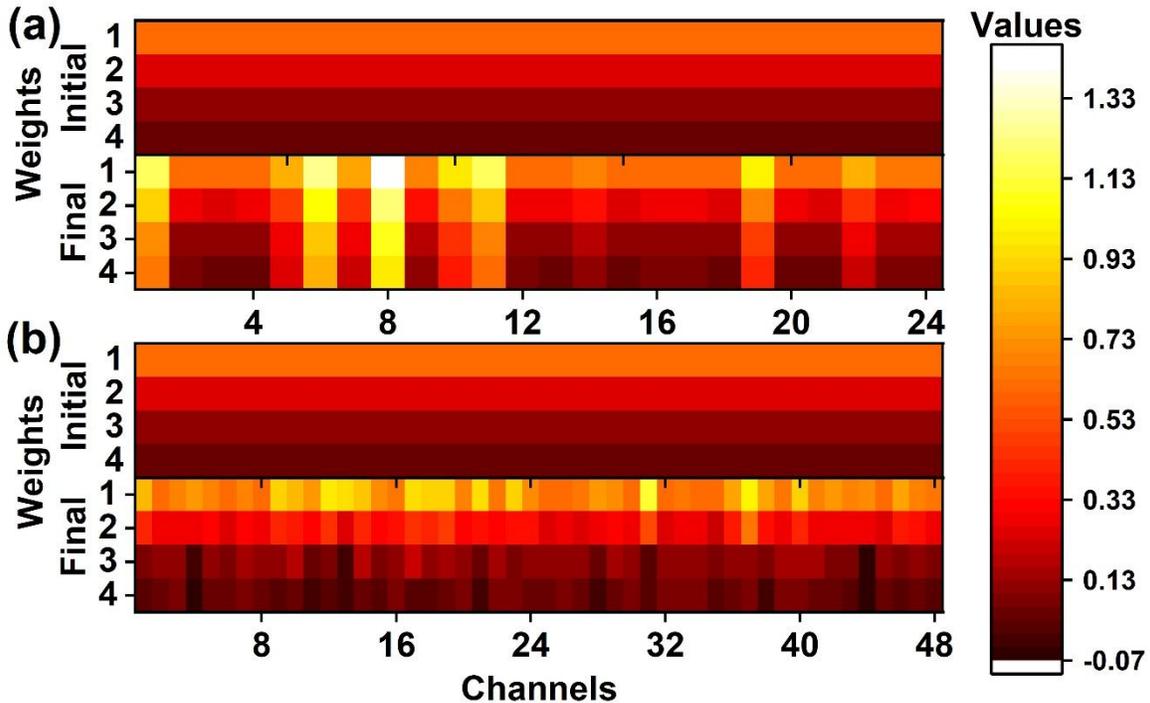

Fig. 4. Heat maps of the weights corresponding to the ordinal pooling networks for different channels at the beginning and after the training for the (a) first and (b) second pooling layers in Fig. 2.

A table summarising the results of validation and test errors is displayed in the inset of Fig. 3. Notice that despite employing a large number of parameters, LBPNs result in higher errors compared to max-pooling on both validation and test sets (each of 10000 images), which is attributed to the absence of nonlinearity in LBPNs. In contrast, OPNs outperform both max-pooling and LBPNs on validation and test sets by at least 0.11% and 0.09%, respectively. Even though these improvement results might not be significant, we expect the performance to scale up in deeper networks that employ numerous layers of pooling. Therefore, as a future work the impact of the proposed scheme will be analysed in a large-scale model on other visual tasks.

## 6. Conclusion

A gradient-based novel pooling scheme, Ordinal Pooling Networks, is introduced in this work, which operate in two steps. In the first step, all the elements of a pooling region are ordered in a sequence. Subsequently, a trainable weight is assigned to each of the elements based upon its order in the sequence. In this work, the activation value of the elements is chosen as a criterion for ordering, however, other criteria can also be envisioned. Our experiments with a small-scale convolutional neural network model upon the image classification task, reveal that replacing max-pooling units with OPNs produces a consistent improvement in the classification accuracy by 0.10%, and also leads to a faster convergence. We expect this performance to improve in the networks employing multiple layers of pooling operations.

## 7. Acknowledgement

The author acknowledges the support from Yulan Liu, Research Scientist at Amazon, for providing the valuable suggestions in improving this article.